# Sorting out symptoms: design and evaluation of the 'babylon check' automated triage system

**Katherine Middleton** BM BCh BA Hons. (Oxon), **Mobasher Butt** BSc (Hons) MB BS (Lon), MRCGP, **Nils Hammerla** Ph.D., **Steven Hamblin** Ph.D., **Karan Mehta** BMedSci BM BS, **Ali Parsa** Ph.D.

babylon, London, UK

**ABSTRACT**
Prior to seeking professional medical care it is increasingly common for patients to use online resources such as automated symptom checkers. Many such systems attempt to provide a differential diagnosis based on the symptoms elucidated from the user, which may lead to anxiety if life or limb-threatening conditions are part of the list, a phenomenon termed 'cyberchondria' [1]. Systems that provide advice on where to seek help, rather than a diagnosis, are equally popular, and in our view provide the most useful information. In this technical report we describe how such a triage system can be modelled computationally, how medical insights can be translated into triage flows, and how such systems can be validated and tested. We present *babylon check*, our commercially deployed automated triage system, as a case study, and illustrate its performance in a large, semi-naturalistic deployment study.

**INTRODUCTION**
In general there are three stages to the healthcare needs of a typical patient. First is the provision of information, wherein a patient may seek to know more about the symptoms or conditions that they have. For example, a patient may seek information about diabetes, or advice on their diet. Secondly, a patient wants to know how to proceed in their care, which is medical triage, and relies on medical training of experienced practitioners. Patients concerned about their symptoms must know if they should go to A&E, book an appointment with their GP, or remain at home and rest. The last stage is that of medical diagnosis, which is by far the most challenging task.

The vast majority of automated systems available online cater to the first need. They provide the digital equivalent of a leaflet at the local surgery, either based on a manual selection (e.g. medical dictionaries), or through a chat-like interaction with an automated agent (e.g. your.md). However, these systems do not provide specific advice on how the user should proceed in their medical care.

Other systems available online allow the user to check their symptoms against a database of known conditions. The automated systems then "diagnose" the conditions that are most likely, given the information provided by the user. This attempt at automated diagnosis can be troublesome for the user, as often potentially life-threatening conditions are reported in long lists of likely candidates. A recent review of automated symptom checkers found that, on average, they only predict the correct diagnosis in 34% of tested cases [2]. However, even if medically correct, reporting these conditions without professional guidance is not in the interest of the user and may lead to increased anxiety [1].

In practice, members of the public often utilize GP consultations simply for reassurance. However, this is not sustainable and leads to over-saturation of GP services [3]. This indicates a need for automated symptom checking systems to give advice on how to resolve symptoms, and reassurance. In other words, patients require systems to cater to their second need: medical triage. There are only few symptom checkers that provide patients with fast and clear advice on what action to take for a particular set of symptoms. The closest type of service that patients can use in the UK is the NHS 111 service, a semi-automated phone-based symptom checker (operated by nurses). The lack of reliable automated triage system is not surprising, as the conception, testing and validation of such services presents a considerable practical challenge.

This paper describes the development process of *babylon check*, a symptom checker that was developed as a collaborative effort between medical experts and engineers, and is deployed as part of our digital health platform. We describe in detail our computational model of triage and how it provides a flexible and efficient basis for modelling triage systems. We illustrate a practical process that can be used to devise the structure of such models and how the medical experts can drive the development process. Furthermore, we describe our testing and validation approach and illustrate the performance of *babylon check* in a large deployment study.

**ONLINE SYMPTOM CHECKERS**
The term "symptom checker" refers to an online tool or application that interprets a user's symptoms in order to provide a likely diagnosis or give guidance on action to take. There are numerous commercial symptom checkers available online. They may be divided into "diagnostic" symptom checkers that provide a list of diagnoses that fit the patient's symptoms, often arranged in order of probability; and "triage" symptom checkers, that do not attempt to provide a diagnosis but direct the patient to the most appropriate source of help for their problem. Diagnostic symptom checkers may be further subdivided into so-called

"specialist" symptom checkers which focus on a particular clinical subspecialty or anatomical region, for example orthopaedics [4], and "generalist" symptom checkers which incorporate a larger range of symptoms.

Symptom checkers are becoming increasingly popular – at its peak, NHS Choices reported 15 million users per month [5], while the commercial symptom checker *iTriage* reports 50 million usages per year [6]. There is, however, little published data regarding the sort of conditions symptom checkers are used for. Similarly, there is currently no standardised assessment criteria for judging the accuracy of a symptom checking tool (although this may change, [7]). To date, the only published study comparing the accuracy of a number of symptom checkers against standard criteria, was completed in [2]. Here, the team interrogated symptom checkers with a series of 45 clinical vignettes with a known diagnosis/urgency. The results showed that, for existing "triage" symptom checkers, accuracy decreased when comparing vignettes requiring emergent intervention, against non-emergent intervention, or simple self-care (80% vs 55% vs 33%).

## REQUIREMENTS

There are a multitude of potential ways to model a triage system. To inform further stages of development a number of requirements were defined in the early stages of the project: clinical safety, utility, speed and flexibility.

### Clinical safety

The triage system is required to hold up to clinical standards for its safety. This includes ensuring that the triage results are reliable and medically correct, but also demands that design decisions are traceable and auditable by authorities. For example, when the user reports symptoms that are consistent with a sprained ankle, and the triage system recommends DIY bed rest, there must be a simple and reliable method for determining why the system made that decision and upon whose authority that advice was given.

### Utility

In order to provide a useful service to the user it is crucial that the triage system is not too risk averse. It is common for automated symptom checkers to avoid potentially risky decisions by reverting to overly pessimistic fallback decisions. For example, one of the triage systems evaluated in [2] always advised the user to seek emergency care, regardless of the type of reported symptoms. This in turn may lead to unnecessary consultations or hospital visits, which is a nuisance to the user and a potential burden to the healthcare system. This is an issue that we aimed to avoid through extensive validation and testing.

### Speed

While it is crucial that all relevant potential symptoms are captured during questioning it is also essential that the process completes in as little time as possible. In practice there is a trade-off between the amount of information gathered from the user and the accuracy and adequacy of the triage decision. However, with increased detail comes increased cost, as users may get irritated by seemingly unrelated questions, and may provide erroneous or even contradicting inputs. Based on our experience in providing clinical consultations we estimated that a typical triage session should take approximately 1-2 minutes, or equal approximately 10-12 questions.

### Flexibility

The last major requirement surrounds the practical constraints of a commercially deployed symptom checker. In daily operation it is essential that the system is easy to extend to new conditions, that it is straight-forward for medical experts to adjust triage decisions, and that altering outcome descriptions and wording is as easy as possible.

## AUTOMATED TRIAGE

When a medical practitioner performs a triage there are two main concerns: i) Does the patient need emergency care (A&E), or ii) what level of non-emergency care is appropriate (GP / Pharmacy / self care). These two concerns require different strategies to perform a successful triage. In order to determine whether emergency care is required the practitioner aims to find certain red flags, which can be individual symptoms (e.g. seizure), or a combination of symptoms (e.g. chest pain and shortness of breath). Red flags are well established and their discovery usually ends a consultation with a reliable outcome of "A&E". The second concern, differentiating non-emergency care, is much more challenging to do reliably. It basically depends on the medical expertise of the practitioner to judge the clinical worry represented by each reported symptom, which is inherently subjective and requires an extensive amount of training.

Automated symptom checkers usually perform poorly when differentiating between non-emergency care settings. For example, [2] found that commercially available online triage-systems only provide the correct triage in 33% of cases for conditions where self-care is appropriate, while they do much better for conditions requiring emergency care (80%).

Computationally, a triage corresponds to a sequence of questions and answers, forming a kind of directed graph. The most straight-forward model would see an outcome, e.g. "stay at home and rest", associated with every possible path in that graph, i.e. to a specific patient history. However, even when limiting the scope of the conditions modelled in the triage it is not practical to construct such a tree, as its size would grow exponentially. Furthermore, it would be difficult to maintain, as large sub-graphs would be replicated along many different paths in that tree.

## MODEL

Based on the requirements set out above we decided on a computational model of triage that can be best described as a directed graph linking a variety of entities, including questions, answers, triggers, and outcomes, which are described below. The graph contains both explicit representation for outcomes to model red flag symptoms as well as machinery for inference based on the weighted symptoms reported by the user. The overall connectivity within the graph is illustrated in Figure 1.

### Questions

Questions are formulated to either check for the presence of a particular event or feature in the patient history ("have you recently injured your head", "do any of these apply to you"), to elucidate specific symptoms ("do you have a fever?") or to further explore a previous response ("How bad is the pain?"). Each question is associated with at least two answers, each of which is potentially linked to a follow-up question. A question that is presented to the user has to receive an answer, and the answers are designed to always include a reject option (e.g. "none of these").

### Answers

Answers can be associated with outcomes, with a score that is added to the overall score if the answer is selected, and with triggers. An answer can further explicitly link to a follow-up question, which will be presented to the user if the answer is selected. The sequence of questions and answers forms the overall structure of the triage flow.

### Triggers

Triggers allow for more sophisticated logic when modelling each triage flow. They represent a logical AND gate that is activated if all linked entities were selected by the user or are themselves activated. Triggers can be conditional on one or more answers to model combination of symptoms seen as red flags. Furthermore, they can be conditional on the overall score (e.g. larger than X), or the age of the user (e.g. younger than X years). Triggers link to outcomes and scores that are only added to the list of selected outcomes or added to the overall score if the trigger condition is satisfied.

### Outcomes

Outcomes link to textual descriptions of advice and fall into a number of severity categories (e.g. "DIY", "HOSPITAL", "GP ACUTE", etc). They are associated with answers either directly (regardless of other answers), or when "triggered" by specified conditions such as duration, low/high score, or a combinations of answers. Outcomes are "collected" during the triage session as the user provides answers. Each outcome has an associated *priority*, allocating a ranking according to urgency e.g. priority 1 is top priority, indicating highest clinical urgency.

### Scores

When a user starts a triage flow the overall score is initialized to zero. Score nodes within the graph will either add or subtract an amount of "points" from the overall score, which is evaluated at the end of the triage to potentially trigger further outcomes. A higher score indicates a higher degree of "clinical worry", or potential for a serious condition requiring clinician assessment. Scores can be linked to answers and to triggers, which may condition a score on a combination of answers. This is to reflect how, in real practice, certain symptoms become more reassuring or more worrying when in combination.

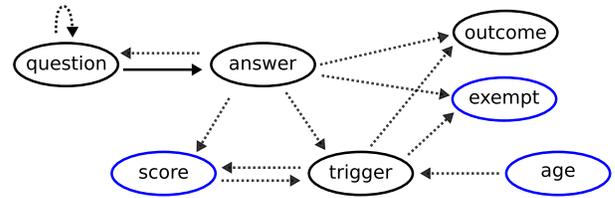

Figure 1: Connectivity in the graph that models a triage flow. Dashed arrows indicate optional links, solid arrows are required. Entities with black outlines form the core structure of the triage graph, while blue entities are used during inference after the session is completed.

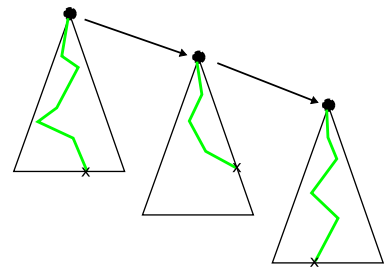

Figure 2: Triage as sequence of trees where each root node is a question (see figure 1). By answering questions, a user traverses each tree until a leaf-node is hit (green lines). This triggers the transitions to the next tree. The triage session stops if all trees have been traversed (see text for details).

### Exempts

Exempts are similar to outcomes and have the same connectivity to other entities in the graph. However, instead of selecting an outcome, they effectively remove this outcome from the final list generated at the end of the triage session. This ensures that the triage will not report a low priority outcome (e.g. "DIY") when red-flag symptoms are selected.

### Outcome selection

In simple terms this triage model represents a sequence of trees, where a specific user session is a path through each tree that always ends in a leaf (see figure 2). There are two procedures to select an outcome for a triage session: i) immediately linked outcomes to red-flag symptoms, and ii) inference based on a weighted scoring of symptoms along with demographic factors.

This corresponds to a separation of the two main concerns during triage described above. On one hand we have well-established medical prior knowledge encoded immediately into the graph structure by linking answers to outcomes. On the other hand, we select outcomes using a weighted scoring scheme to model non-emergency outcomes.

**DESIGN PROCEDURE**

This section gives a brief overview of the procedure used to construct the triage flows used in *babylon check*. For each body part, a list of common or important pathologies was generated, including e.g. trauma, infection, emergencies, and neurological disorders. Based on expert medical opinion we grouped the pathologies by most appropriate triage-outcome. For each pathology a number of key factors were identified that separate the condition (e.g. fever, shortness of breath, etc.).

**Overall structure and emergency care**

The basic structure of the triage flow was established following a procedure similar to the training of decision trees [8]. Medical experts selected the differentiating factor that best separated the conditions requiring emergency care, and added a corresponding question (e.g. "Have you recently injured your arm?"). We continued to add questions while it was medically meaningful to separate the pathologies based on the most informative factors. For each of the questions identified in this process we added additional "child" questions that separated the conditions within that category (e.g. "Are you in pain?") following the same process on a subset of pathologies. After this process was complete, and based on well-established medical knowledge, we associated outcomes to red flag symptoms or groups of symptoms through the use of triggers.

**Non-emergency care**

Reliably differentiating non-emergency care paths goes beyond the structural representation used for emergency care. Instead, it requires an inference procedure that models the degree of "clinical worry" associated with different symptoms and groups of symptoms. The process we devised is initially based on the intuition of medical experts, which is used to bootstrap an iterative optimization scheme. During this optimization we continuously improve the performance of *babylon check* on conditions that do not require emergency care. A detailed description of the approach goes beyond the scope of this paper.

**Summary**

Our symptom checker requires a user to select a body part and answer a series of multiple choice questions. The engine "collects" outcomes according to which answers are selected, and whether the triggers for those outcomes are satisfied. The engine discards outcomes according to presence of exemptions. The result is a list of possible outcomes, of which the highest priority outcome is chosen and presented to the user.

**ITERATIVE VALIDATION PROCEDURE**

*babylon check* was tested by experienced clinicians both from within the babylon clinical development team as well as outside consultants to ensure unbiased and reliable assessment. We opted for an iterative validation approach that consists of multiple cycles of expert review and consensus-based improvements to the design of *babylon check*. Each iteration is summarized below.

**First stage**

A group of primary care physicians reviewed all triage flows established with the procedure outlined above, and independently trialed all medical conditions that were modelled during that process. They compared their expected outcome with the outcome produced by *babylon check*. They were also encouraged to model (i.e. formalize for use in the development of check) any other condition they commonly saw in their everyday practice or deemed clinically important. Based on feedback by the clinicians and captured data points the babylon core clinical development team (CCD) decided which changes to implement based on a consensus approach.

**Second stage**

A large selection of medical experts (primary care physicians, external emergency medicine consultants, and external pharmacists) conducted the second stage review. Each primary care physician modelled all original conditions used in the design process, and any additional conditions added in the first review stage, based on their experience and clinical knowledge. Comparison was made between their expected outcome for a condition, and the outcome received from *babylon check*. This stage was complemented with a further language review of outcome screens, questions, and answers to ensure clinical accuracy and patient friendly terminology. Emergency medicine consultants reviewed all hospital triggers. External pharmacists were consulted in a workshop format to give feedback regarding the pharmacy outcome screens and the conditions sent to pharmacy in all stages of development. All clinical scenarios and feedback were captured for future use, and informed further changes by the CCD team.

**Third stage**

A large group of clinicians (babylon primary care physicians, lead babylon specialists, and external doctors) conducted a formal testing exercise for a number of conditions of their own choice with no pre-defined scenarios. Feedback from this stage again informed changes to *babylon check* by the CCD team.

**Evaluation**

The babylon CCD team used a total of 33 clinical scenarios validated by external specialists [2] to test *babylon check*. These standardized patient vignettes were equally divided into three categories of triage urgency: emergent care required (for example, pulmonary embolism), non-emergent care reasonable (for example, otitis media), and self care reasonable (for example, viral upper respiratory tract infection). Performance during this evaluation further informed final reviews to *babylon* check by the CCD team.

We found our approach to work significantly better than the average performance across automated triage systems reported in [2]. Particularly the performance for non-emergency care and self-care, by far the dominant type of request to such systems, exceeds that of comparable systems significantly. It is important to stress that the "correct" outcome for a patient vignette may depend on the healthcare system it is aimed at. As only a subset of vignettes from [2] is applicable to our system (33/45, pediatrics still in development), we opted to postpone the presentation of these results to a later version of this manuscript.

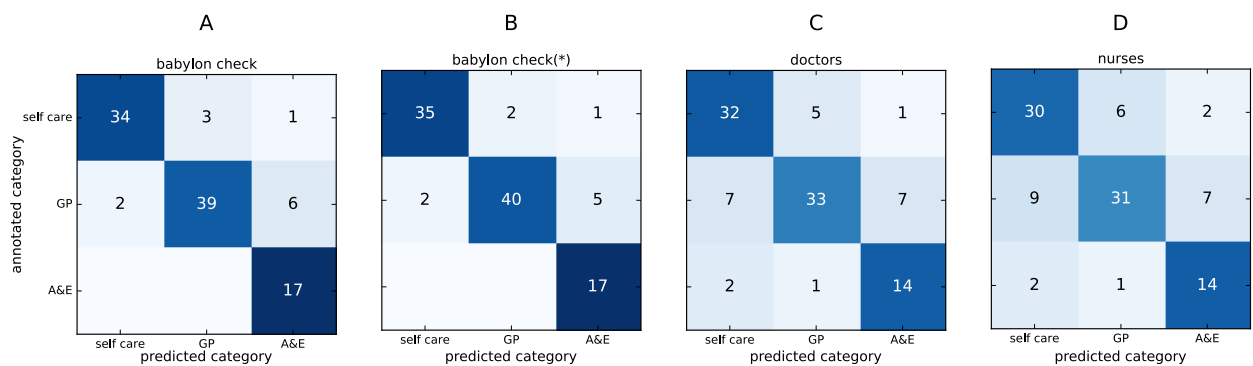

Figure 3: Confusion matrices for each approach, illustrating the performance across the patient vignettes in the deployment study (n=102). Rows contain vignettes annotated for each category, columns indicate their predicted category. (*) indicates ideal input to *babylon check* (see text for details).

**DEPLOYMENT STUDY**

With the iterative design process completed we aimed to demonstrate the clinical validity of *babylon check* at a level consistent with the requirements for typical medical devices. In order to confirm that *babylon check* performs in line with the requirements specified at the beginning of this document a deployment study was performed.

**Outline**

We recruited 12 clinicians that had at least 4 years of clinical experience, along with 17 nurses, to perform triage in a semi-naturalistic scenario. We further recruited professional actors that, during the study, had mock consultations based on patient vignettes with both clinicians and nurses. Finally, the actors also used the *babylon check* app where the automatically generated outcome was captured for later comparison. All consultations and the use of the app were timed.

**Patient vignettes**

A medical expert employed by babylon - but whom had not been involved in the development of *babylon check* - devised a total of 102 patient vignettes, each containing a detailed description of symptoms for a condition. These vignettes were designated as falling within one of 5 outcomes of varying urgency: *A&E, GP urgent, GP routine, Pharmacy*, and *Manage at home*. All vignettes were reviewed by independent experts for clarity and accuracy of the associated triage outcome. A gold standard outcome was also obtained for each vignette through a majority vote on the independent ratings of three expert clinicians.

**Process**

The study was conducted over a period of approx. two weeks. Each actor was instructed to answer questions during consultations solely based on the information provided on the vignette. In case of doubt or if the answer was not listed on the vignette the actor consulted the supervising clinician. Actors were further advised to try to emulate a realistic consultation and to not provide the information from the vignette unless it was asked for during the consultation. Actors were allocated a new vignette at random before each consultation and given time to study it and have questions clarified. No actor used the same vignette twice (e.g. for doctor's consultation and nurse's consultation). Finally, the actors ran through the triage process using the *babylon check* app.

**Results**

As is common practice we grouped the outcomes for each vignette into three categories [2]: self-care (self-management / pharmacy), GP, and emergency care (A&E). Confusion matrices illustrating the performance of each approach are shown in figure 3. Overall we see that *babylon check* outperforms both doctors and nurses: an accurate outcome (i.e. at the same urgency) is produced in 88.2% of cases for *babylon check*, in 75.5% of cases for doctors, and in 73.5% of cases for nurses. In order to exclude input errors performed by the actors when using *babylon check* we further had a medical expert perform one run of triage through *babylon check* for each vignette. This slightly improved the performance of check, where 90.2% of vignettes were classified correctly (see figure 3B). We further observe that, in this study, there was no under-triaging of emergency cases, which all received an appropriate emergency outcome from *babylon check* (see figure 3A/B).

Our system effectively performs triage of both emergency cases as well as non-emergency cases. Interestingly these categories show different characteristics in their prediction: it is crucial for emergency cases to receive an appropriate, potentially life-saving outcome. Our approach enables the reliable use of red-flags which leads to a high recall of 100% (all cases needing A&E will receive that outcome), at the cost of precision (some cases falsely receive A&E outcome), as shown in table 1. The optimization of the triage outcomes for non-emergency cases is, however, not driven by red-flags, but based on a more sophisticated inference procedure. This enables prediction with high precision at the cost of recall (potentially over-triaging for border-line cases).

Table 1: Performance of all approaches across severity-categories: Recall (R) and Precision (P), overall accuracy, and time taken to triage. On average, *babylon check* outperforms both doctors and nurses in this study, being more accurate and requiring significantly less time. (*) indicates ideal input to check.

|  | Nurses | | Doctors | | Check | |
| --- | --- | --- | --- | --- | --- | --- |
|  | R | P | R | P | R | P |
| Self-care | 0.789 | 0.732 | 0.842 | 0.780 | 0.895 | 0.944 |
| GP | 0.660 | 0.816 | 0.702 | 0.846 | 0.830 | 0.929 |
| A&E | 0.824 | 0.609 | 0.824 | 0.636 | 1.000 | 0.708 |
| Mean | 0.758 | 0.719 | 0.789 | 0.754 | **0.901** | **0.860** |
| Accuracy | 73.5% | | 77.5% | | **88.2%** | |
| (*) | - | | - | | **90.2%** | |
| Time | 2:27 (±1:03) | | 3:12 (±1:08) | | **1:07 (±0:27)** | |

Additionally, a review of clinical safety was conducted for all the outcomes obtained through *babylon check* (actors), as well as for doctors and nurses (see figure 4). All outcomes produced by our system for the 102 patient vignettes were clinically safe (two cases of under-triaging were medically reasonable to triage to self-care). Of those cases under-triaged by nurses, internal review found three to be medically unsafe (i.e. should have received emergency care but did not), corresponding to 97% clinical safety. The doctors in our study failed to associate two vignettes to A&E, leading to 98% clinical safety (albeit on different cases to the nurses). This unexpected lack of clinical safety can potentially be attributed to actors deviating from the patient vignettes during mock consultations, an issue we are currently investigating in our continued validation and testing efforts.

All the triage session were timed in this study. *babylon check* was fastest in 89% of cases (91/102), followed by the nurses with 9% (9/102) and the doctors at 2% (2/102). On average a consultation with a doctor took 03:12 (+/-01:08), with a nurse 02:27 (+/- 01:03). Consultations with *babylon check* took on average only 01:07 (+/- 00:27).

In this study, a one-tailed t-test indicates that consultation time of check was significantly faster than doctors (p<0.001) and nurses (p<0.001). These results are very promising and indicate the immense potential for automated triaging systems as a practical route to cost savings in healthcare systems.

## SUMMARY
In this technical report we highlighted and described the development of *babylon check*, the automatic triage system deployed in our digital health platform. We discussed in detail how triage systems can be modelled and described the incremental testing and validation process that leads to a clinically safe and efficient triage system.

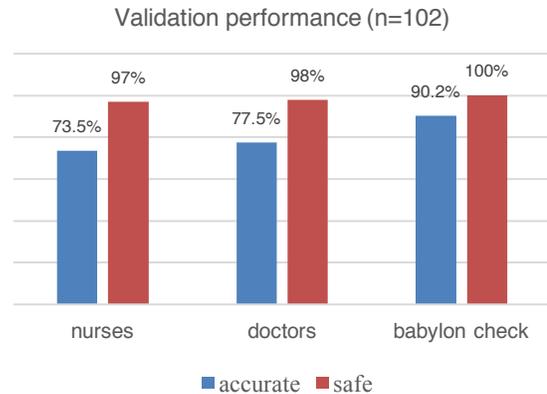

Figure 4: Validation performance in deployment study.

In a semi-naturalistic study we found that our system outperforms both doctors and nurses across 102 mock patient consultations. *babylon check* provided a clinically safe outcome in 100% of cases, and performed an accurate triage in up to 90.2% of cases. Further it required the least amount of time in 89% of all consultations, effectively cutting the time to triage by half for the average case. Our results indicate that automated systems have a promising future in medical triage with the potential to reduce cost and improve access to healthcare.